\documentclass{article}
\usepackage{iclr2027,times}

\usepackage{amsmath,amsfonts,bm}

\def\eqref#1{equation~\ref{#1}}

\def\1{\bm{1}}

\DeclareMathAlphabet{\mathsfit}{\encodingdefault}{\sfdefault}{m}{sl}
\SetMathAlphabet{\mathsfit}{bold}{\encodingdefault}{\sfdefault}{bx}{n}

\usepackage{amsmath}
\usepackage{amssymb}
\usepackage{booktabs}
\usepackage{graphicx}
\usepackage{wrapfig}
\usepackage{multirow}
\usepackage{makecell}
\usepackage{tabularx}
\usepackage{capt-of}
\usepackage{enumitem}
\usepackage[table]{xcolor}
\usepackage{hyperref}
\usepackage{url}

\definecolor{best}{RGB}{255, 180, 180}
\definecolor{second}{RGB}{255, 230, 180}
\definecolor{third}{RGB}{255, 255, 200}
\definecolor{placeholdergray}{RGB}{245,245,245}
\title{InfraWorld: Motion-Aware World Modeling for Roadside 4D Occupancy Forecasting}

\author{
Songkai Wang\textsuperscript{1}\textsuperscript{*},
Xiaokai Bai\textsuperscript{1}\textsuperscript{*}\textsuperscript{$\dagger$},
Lei Yang\textsuperscript{2},
Hui-liang Shen\textsuperscript{1}
\\[2mm]
\textsuperscript{1}College of Information Science and Electronic Engineering, Zhejiang University.
\\
\textsuperscript{2}School of Mechanical and Aerospace Engineering, Nanyang Technological University.
\\
\textsuperscript{*} Equal contribution, \textsuperscript{$\dagger$} Corresponding author.
\\
\texttt{email:songkaiw0805@gmail.com}
}

\iclrfinalcopy
\begin{document}

\maketitle

\begin{abstract}
Roadside 4D occupancy forecasting predicts future semantic occupancy in a fixed infrastructure coordinate system. Roadside scenes are globally structured yet locally evolving, requiring comprehensive scene-level context while precisely modeling sparse localized motion. However, existing forecasting paradigms struggle to tackle both requirements. To address this limitation, we propose \textbf{InfraWorld}, a world model formulating roadside occupancy forecasting as selective state evolution. InfraWorld maintains a dense BEV state for the complete scene and persistent motion queries for moving regions, assigning complementary states to global context and local dynamics. Specifically, we propose Dynamic Query Initiation (DQI) to initialize queries in regions supported by motion in the observed history. During rollout, the dense state is updated through Long--Short Temporal Fusion (LST), which combines short-term adjacent-frame continuity with long-term motion trends to preserve stable context while capturing scene evolution. Motion-Aware Query Injection (MAQI) then refines the queries and injects confidence-weighted query information into corresponding BEV tokens, enabling targeted dynamic updates without perturbing global context. Together, InfraWorld concentrates predictive capacity on evolving traffic regions. Extensive experiments on the InfraOcc and Occ3D--nus datasets demonstrate that InfraWorld achieves state-of-the-art performance for roadside 4D occupancy forecasting while retaining its effectiveness under ego-centric settings. Code will be released.

\end{abstract}

\section{Introduction}

Semantic occupancy represents free space, static infrastructure, and traffic participants in a unified 3D field~\citep{MonoScene,VoxFormer,TPVFormer,OccFormer,uniscene}. Extending this representation over time, 4D occupancy forecasting predicts future semantic occupancy from observed sequences~\citep{Cam4DOcc,OccWorld,OccProphet,I2World} for
planning, safety assessment, and simulation. However, existing studies have primarily focused on ego-centric settings. In contrast, roadside perception continuously observes a fixed traffic space with infrastructure-mounted sensors, often beyond the view of individual vehicles, supporting
traffic monitoring and coordinated safety services.~\citep{Rope3D,DAIRV2X,V2XReal,RCooper,TUMTrafV2X}. Recently, InfraOcc~\citep{InfraOcc} established fixed-coordinate 3D occupancy prediction for roadside scenes, yet roadside 4D occupancy forecasting remains unexplored. 

\begin{figure*}[t]
    \centering
    \includegraphics[width=\textwidth]{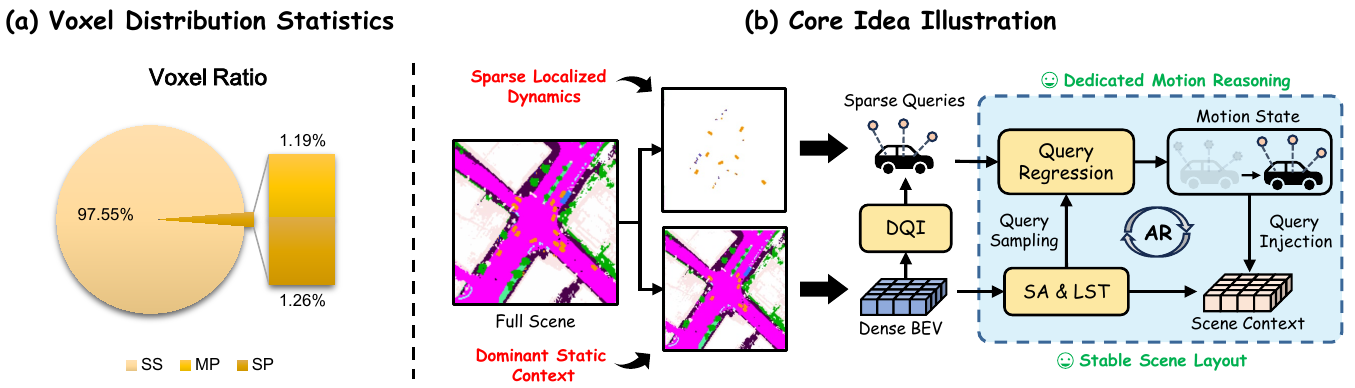}
    \caption{(a) Voxel distribution in InfraOcc: static scene (SS), moving participants (MP), and stationary participants (SP). (b) InfraWorld maintains the stable scene layout with a dense BEV state, while queries reason about localized dynamics and inject their updates into the scene context.}
    \label{fig:insight}
    \vspace{-18pt}
\end{figure*}

The fixed viewpoint gives roadside occupancy forecasting a distinctive spatiotemporal structure. As shown in Fig.~\ref{fig:insight}(a), static infrastructure and stationary elements dominate the scene and remain spatially aligned over time, while moving participants occupy only a few localized regions yet account for most future changes. Consequently, roadside occupancy forecasting is globally structured yet locally evolving. On the one hand, forecasting a complete occupancy field requires scene-level context to preserve the overall spatial layout and provide a coherent global scaffold for future prediction. On the other hand, the actual scene evolution is determined by sparse local regions, whose continuous motion requires fine-grained and flexible modeling of their spatial displacement and temporal dynamics. An effective roadside world model therefore needs to retain comprehensive scene-level context while precisely modeling the sparse local motions that drive future changes, which raises a key question: \textit{how should a world model reason when scene state and scene evolution have fundamentally different spatial support?}

Existing ego-centric forecasting paradigms address this asymmetry only partially. Dense BEV-based methods ~\citep{OccWorld,I2World,DriveOccWorld} maintain a complete scene representation and effectively capture scene-level context, but their fixed-resolution grids evolve the scene uniformly, allowing dominant unchanged regions to dilute the sparse changes that drive future evolution. Flow-based methods ~\citep{DFITOccWorld,FSFNet} explicitly model such changes by predicting displacement and warping occupancy or features, but primarily formulate scene evolution as spatial transport, which limits their ability to flexibly reason about the continuous evolution of localized traffic regions. Thus, existing formulations still struggle to preserve comprehensive scene context while devoting sufficient modeling capacity to the sparse local dynamics that determine future changes.

To reconcile comprehensive scene modeling with fine-grained reasoning over localized dynamics, we formulate roadside occupancy forecasting as selective state evolution, where persistent scene context and sparse scene changes are assigned complementary modeling roles. We instantiate this principle in \textbf{InfraWorld}, a roadside occupancy world model that maintains a dense BEV state for the spatially complete scene context while using a small set of persistent motion queries that flexibly reason about localized traffic evolution, as illustrated in Fig.~\ref{fig:insight}(b). InfraWorld first employs Dynamic Query Initiation (DQI) to initialize queries where motion is likely to occur with motion evidence from the observed history. For temporal fusion, we develop Long--Short Temporal Fusion (LST) which combines recent continuity with longer-term change trends to model temporal evolution at complementary scales. Motion-Aware Query Injection (MAQI) then transfers reliable query-level dynamics to nearby BEV tokens, strengthening evolving regions without disrupting the stable scene context. With these designs, InfraWorld preserves complete roadside structure while focusing predictive capacity on localized dynamics. Our contributions are summarized below.
\begin{itemize}[leftmargin=*,itemsep=2pt,topsep=2pt]
    \item We propose \textbf{InfraWorld}, the first framework for roadside 4D occupancy forecasting to the best of our knowledge, which formulates roadside forecasting as selective state evolution and assigns complementary modeling roles to spatially complete scene context and sparse localized dynamics.
    \item We introduce Dynamic Query Initiation (DQI) and Motion-Aware Query Injection (MAQI) to allocate dedicated motion reasoning to regions likely to evolve and selectively transfer reliable dynamic information back to the scene state.
    \item We develop Long--Short Temporal Fusion (LST), which combines recent-frame continuity with longer-term trends to strengthen temporal reasoning for scene evolution during rollout.
    \item Extensive experiments on InfraOcc and Occ3D--nus validate the effectiveness of InfraWorld for roadside 4D occupancy forecasting and its applicability to ego-centric settings.
\end{itemize}

\section{Related Work}
\label{sec:related_work}

\subsection{4D Occupancy World Models}
\label{sec:rw_world_models}

4D occupancy world models forecast future semantic occupancy from observed scene history. Early methods predict LiDAR-derived occupancy through differentiable ray casting or point-cloud forecasting~\citep{Khurana2022,PointCloudForecasting}, while Cam4DOcc~\citep{Cam4DOcc} introduced a camera-only benchmark and evaluation protocol. Subsequent approaches evolve compact latent scene representations. OccWorld~\citep{OccWorld} combines VQ-VAE tokenization with autoregressive scene-token prediction, whereas RenderWorld~\citep{RenderWorld} separately models empty and occupied voxels in latent space. Other studies incorporate textual scene knowledge~\citep{OccLLaMA,OccLLM}, exploit 4D occupancy reconstruction for representation learning~\citep{DriveWorld,UniWorld}, or condition generation on ego actions~\citep{DriveOccWorld}. For long-horizon forecasting, OccSora~\citep{OccSora} and DOME~\citep{DOME} adopt diffusion-based generation, while I$^2$-World~\citep{I2World} decouples spatial compression from temporal modeling. These methods primarily target ego-centric scenes, where ego motion and traffic dynamics jointly affect the global scene state. In contrast, InfraWorld operates in fixed roadside coordinates and assigns complementary roles to stable global context and localized dynamics.

\subsection{Infrastructure-Side Occupancy and Motion-Aware Scene Modeling}
\label{sec:rw_roadside_motion}

\begin{wrapfigure}[14]{r}{0.5\linewidth}
    \setlength{\intextsep}{0pt}
    \vspace{-1.5\baselineskip}
    \centering
    \includegraphics[width=\linewidth]{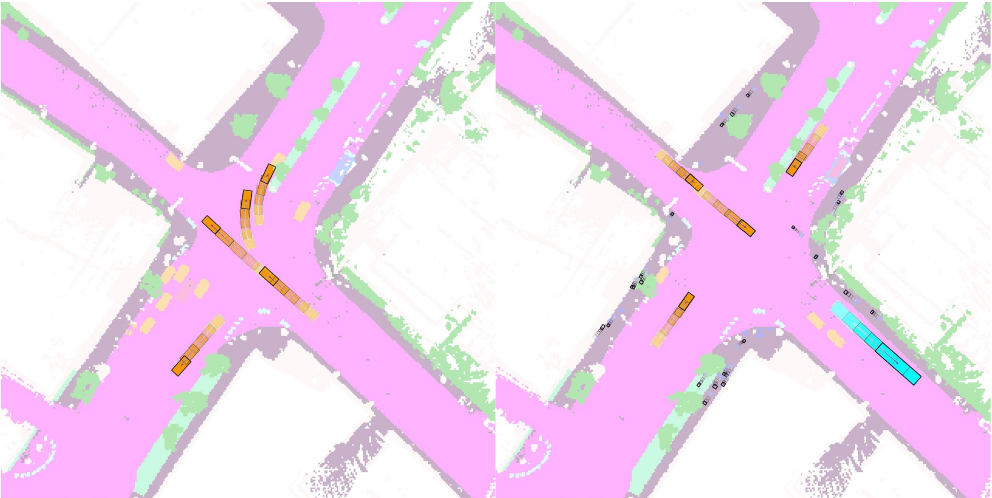}
    \caption{Sparse and localized dynamics in roadside occupancy. Across five consecutive frames, the static scene layout remains largely unchanged, while only a few moving traffic participants exhibit temporal changes.}
    \label{fig:sparse_dynamics}
\end{wrapfigure}
Infrastructure-side occupancy forecasting operates in a fixed roadside coordinate system, where stationary sensors keep roads, buildings, and other infrastructure spatially aligned over time. As illustrated in Fig.~\ref{fig:sparse_dynamics}, this stable background remains largely unchanged across consecutive occupancy frames, while temporal changes are concentrated in a small number of localized moving traffic participants. InfraOcc~\citep{InfraOcc} formalizes this setting and highlights its static--dynamic asymmetry. Roadside forecasting must therefore preserve a spatially complete, temporally stable scene scaffold while focusing modeling capacity on the few regions that undergo change.

Related motion-aware occupancy forecasting methods follow two main paradigms. Flow-based approaches explicitly model displacement: DFIT-OccWorld~\citep{DFITOccWorld} decouples static and dynamic voxels and warps dynamic occupancy with predicted flow, while FSF-Net~\citep{FSFNet} combines coarse BEV-flow forecasting with latent-state prediction. Query-based approaches ~\citep{SparseWorld} allocate computation selectively, representing scene evolution using adaptive sparse queries. Although developed for ego-centric scenes, these paradigms have complementary limitations for roadside forecasting. Flow-based methods primarily treat motion as spatial transport without maintaining a feature-rich persistent state for each moving region, whereas fully sparse query representations are less suited to preserving the spatially complete roadside background. InfraWorld addresses this trade-off by using dense BEV latents for global scene context and persistent motion-focused queries for localized dynamics.

\section{Method}
\label{sec:method}

\subsection{Overview}
\label{sec:method_overview}

Given a sequence of $H$ semantic occupancy observations
$\mathcal{O}_{t-H+1:t}=\{O_{t-H+1},\ldots,O_t\}$ in a fixed roadside
coordinate system, InfraWorld predicts the subsequent $F$ occupancy volumes:
\begin{equation}
    \hat{\mathcal{O}}_{t+1:t+F}
    =\mathcal{F}_{\theta}(\mathcal{O}_{t-H+1:t}),
    \label{eq:forecasting_problem}
\end{equation}
where $\theta$ denotes the model parameters. A pretrained and frozen VQ
tokenizer $(\mathcal E,\mathcal D)$ encodes $O_s$ into a BEV latent
$Z_s=\mathcal E(O_s)\in\mathbb R^{C\times H_z\times W_z}$ and decodes each
forecast latent as $\hat O_s=\mathcal D(\hat Z_s)$.

InfraWorld autoregressively propagates a dense latent $\hat Z_\tau$ for the
complete scene and a sparse query set $Q_\tau$ for predicted moving regions.
Let $\mathcal H_\tau=[\mathbf H_\tau^{(1)},\ldots,
\mathbf H_\tau^{(L)}]$ denote a first-in--first-out queue of $L$ dense
latents ordered from oldest to newest. We initialize it from the $L$ most
recent observations as $\mathcal H_t=[Z_{t-L+1},\ldots,Z_t]$, where
$L\leq H$. After predicting $\hat Z_\tau$, we update the queue by
\begin{wrapfigure}[29]{r}{0.45\linewidth}
    \setlength{\intextsep}{0pt}
    \centering
    \includegraphics[width=\linewidth]{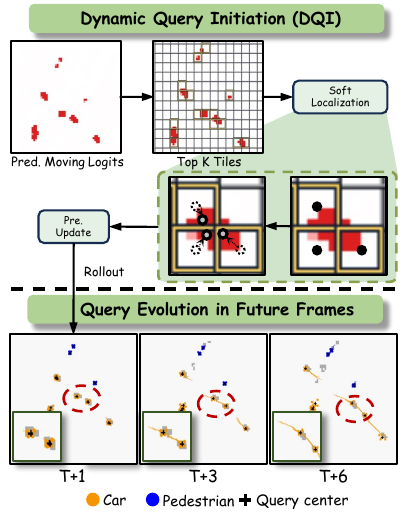}
    \caption{Dynamic Query Initiation (DQI). Top: DQI process for selecting and localizing motion-supported queries. Bottom: query evolution during rollout, where gray regions denote GT moving objects and highlighted trajectories show how queries capture localized motion evolution.}
    \label{fig:query_DQI}
\end{wrapfigure}
\begin{equation}
    \mathcal H_\tau=[\mathbf H_{\tau-1}^{(2)},\ldots,
    \mathbf H_{\tau-1}^{(L)},\hat Z_\tau].
    \label{eq:history_update}
\end{equation}
With $\hat Z_t=Z_t$ and $Q_t$ initialized by Dynamic Query Initiation
(DQI; Section~\ref{sec:query_init}), for $\tau=t+1,\ldots,t+F$, each forecasting step is
\begin{equation}
    (\hat Z_{\tau},Q_{\tau})=
    \mathcal{W}_{\theta}
    (\hat Z_{\tau-1},\mathcal{H}_{\tau-1},Q_{\tau-1}).
    \label{eq:hybrid_transition}
\end{equation}

\begin{figure*}[t]
    \centering
    \includegraphics[width=\textwidth]{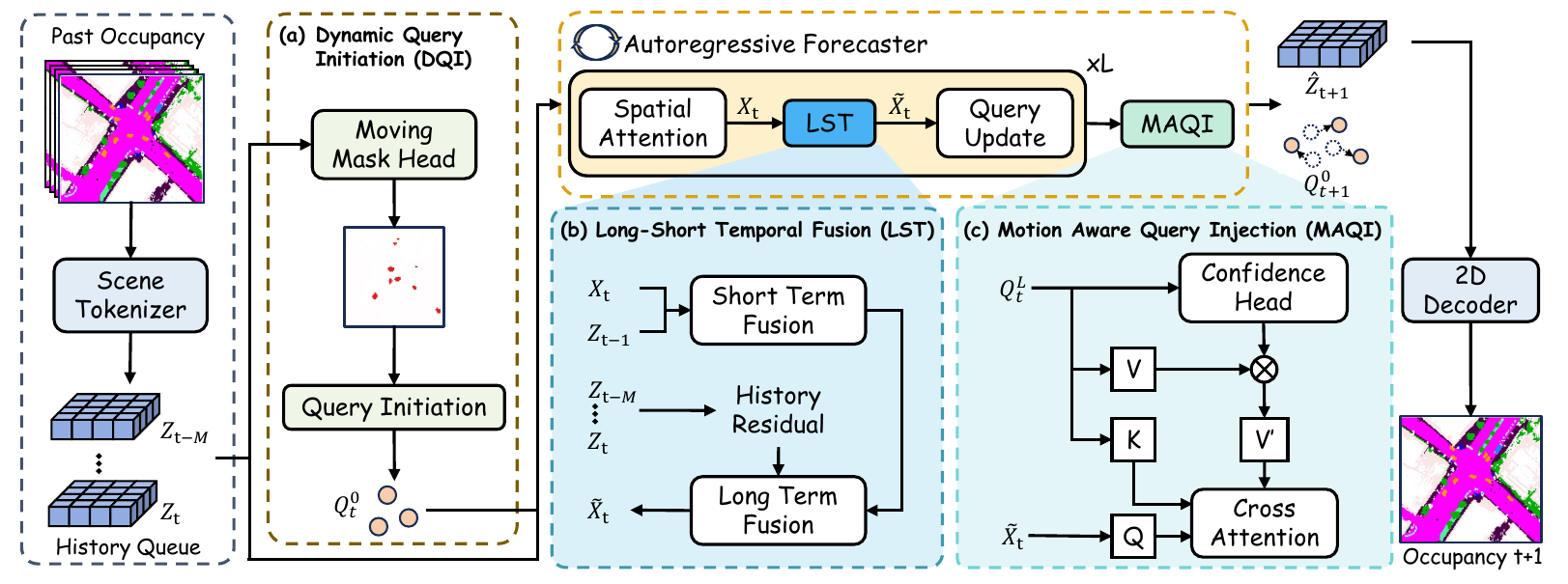}
    \caption{Architecture overview of InfraWorld. A frozen tokenizer encodes the observed occupancy into dense BEV latents, while DQI initializes persistent queries in predicted moving regions. At each autoregressive step, LST fuses spatial and temporal context to update the dense state and refine the queries, after which MAQI injects reliable query features into local BEV regions. The updated dense latent and queries are then propagated to the next forecasting step.}
    \label{fig:architecture}
    \vspace{-18pt}
\end{figure*}

As shown in Figure~\ref{fig:architecture}, $\mathcal W_\theta$ extracts
multi-scale BEV features from $\hat Z_{\tau-1}$ and fuses them with
$\mathcal H_{\tau-1}$ through Long--Short Temporal Fusion (LST;
Section~\ref{sec:temporal_refinement}). The fused features recurrently refine
the queries, after which Motion-Aware Query Injection (MAQI;
Section~\ref{sec:interactive_decoder}) writes reliable query information into
local BEV regions. A residual predictor then updates
$\hat Z_\tau$ from $\hat Z_{\tau-1}$, and the new latent is appended to the
history queue according to Eq.~\ref{eq:history_update}.

\subsection{Dynamic Query Initiation}
\label{sec:query_init}

DQI initializes a limited set of persistent queries in motion-supported regions
to provide dedicated representations for localized dynamics. Given the observed
BEV latents $\{Z_s\}_{s=t-H+1}^{t}$, the query state propagated at rollout step
$\tau$ is defined as
\begin{equation}
Q_{\tau}=\{(\mathbf f_i^{\tau},\mathbf P_i^{\tau},m_i)\}_{i=1}^{N_q},
\label{eq:query_state}
\end{equation}
where $N_q$ is the fixed query budget,
$\mathbf f_i^{\tau}\in\mathbb R^C$ denotes the query feature,
$\mathbf P_i^{\tau}\in[0,1]^{N_p\times2}$ contains $N_p$ normalized point
coordinates, and $m_i\in\{0,1\}$ is the validity mask determined at
initialization. The feature $\mathbf f_i^{\tau}$ encodes regional semantic and
motion information, while $\mathbf P_i^{\tau}$ specifies where the query samples
and updates local BEV features.

A lightweight moving-mask head fuses the current latent with the
observed history to predict moving logits
$\ell\in\mathbb{R}^{H_m\times W_m}$. As illustrated in
Figure~\ref{fig:query_DQI}, we partition $\ell$ into non-overlapping tiles,
rank them by their max-pooled logits, and select the $N_q$ highest-scoring
tiles.

For each selected tile $\mathcal T_i$, its anchor is obtained by soft
localization:
\begin{equation}
    \mathbf a_i = \sum_{u\in\mathcal T_i}
    \frac{\exp(\ell_u/T)}
         {\sum_{v\in\mathcal T_i}\exp(\ell_v/T)}
         \,\mathbf x_u,
    \label{eq:query_evolution}
\end{equation}
where $\mathbf x_u$ is the normalized center of cell $u$ and $T>0$ is the
temperature. A query is marked invalid if the mean moving probability of its
tile falls below a threshold. Tile-wise selection distributes the limited query
budget across candidate motion regions, while soft localization places each
anchor according to the intra-tile motion evidence.

For each valid query, the initial feature is formed by adding a learned query
embedding to the current latent feature bilinearly sampled at $\mathbf a_i$,
and is then refined using the observed history. Following
SparseWorld~\citep{SparseWorld}, the coarse-to-fine refinement samples local
history features around the query points, performs query-conditioned mixing
across sampled points and channels, exchanges information among queries through
self-attention, and predicts point semantics together with an expanded point
set for the next stage. Invalid queries retain their states but are masked from
query interactions. The resulting queries constitute the persistent motion
states that are recurrently updated during forecasting.

\subsection{Long--Short Temporal Fusion}
\label{sec:temporal_refinement}

LST is designed to jointly capture immediate temporal continuity and accumulated
motion trends during autoregressive forecasting. At rollout step $\tau$, the
dense BEV backbone progressively applies spatial self-attention to
$\hat Z_{\tau-1}$ at multiple BEV resolutions. The resulting multi-resolution
features are then upsampled and fused to obtain the BEV feature $\mathbf X_\tau$.
 
For short-term temporal modeling, let $\bar{\mathbf H}_{\tau-1}^{(L-1)}$ denote
the preceding history feature with its temporal embedding. Since the roadside
coordinate system remains fixed over time, it can be directly fused with
$\mathbf X_\tau$ through a ConvGRU-style update:
\begin{align}
    \mathbf z_\tau &= \sigma(\operatorname{Conv}_z
        ([\mathbf X_\tau,\bar{\mathbf H}_{\tau-1}^{(L-1)}])), \\
    \mathbf r_\tau &= \sigma(\operatorname{Conv}_r
        ([\mathbf X_\tau,\bar{\mathbf H}_{\tau-1}^{(L-1)}])), \\
    \mathbf S_\tau &= \operatorname{Conv}_o\!\left(
        (1-\mathbf z_\tau)\odot\mathbf X_\tau
        +\mathbf z_\tau\odot
        \phi(\operatorname{Conv}_h
        ([\mathbf X_\tau,
        \mathbf r_\tau\odot\bar{\mathbf H}_{\tau-1}^{(L-1)}]))
        \right).
    \label{eq:short_fusion}
\end{align}
Here, $[\cdot,\cdot]$ denotes channel-wise concatenation, $\odot$ denotes
element-wise multiplication, and $\phi$ is the candidate activation function.
The update gate adaptively balances the current feature and history-conditioned
information, while the reset gate suppresses irrelevant historical content,
yielding the short-term state $\mathbf S_\tau$.

To capture longer-term motion trends, the long-term branch operates on the raw
history queue $\mathcal H_{\tau-1}$. It first computes adjacent residuals
$\Delta\mathbf H_{\tau-1}^{(j)}=\mathbf H_{\tau-1}^{(j+1)}-
\mathbf H_{\tau-1}^{(j)}$ for $j=1,\ldots,L-1$. A temporal Conv1D is then
applied independently at each BEV location, followed by temporal averaging and
a $3\times3$ spatial aggregation layer to produce the trend feature
$\mathbf R_\tau$. This feature refines the short-term state as
\begin{equation}
    \widetilde{\mathbf X}_\tau =
    \mathbf S_\tau + \mathcal{P}_l([\mathbf S_\tau,\mathbf R_\tau]),
    \label{eq:long_short}
\end{equation}
where $\mathcal{P}_l$ denotes a lightweight projection whose final layer is
zero-initialized. The refined feature $\widetilde{\mathbf X}_\tau$ is
subsequently used to update the dynamic queries and predict the next dense latent
state. In this way, LST combines local temporal continuity with accumulated
motion trends to provide a temporally informed dense representation for both
query refinement and global scene prediction.
 
\subsection{Motion-Aware Query Injection}
\label{sec:interactive_decoder}

MAQI transfers query-level motion information to local BEV regions while
suppressing unreliable updates during autoregressive rollout. At each
forecasting step, the queries are recurrently refined from the current BEV
features with a fixed number of query points.

\begin{wrapfigure}[17]{r}{0.45\linewidth}
    \setlength{\intextsep}{0pt}
    \vspace{-1.5\baselineskip}
    \centering
    \includegraphics[width=\linewidth]{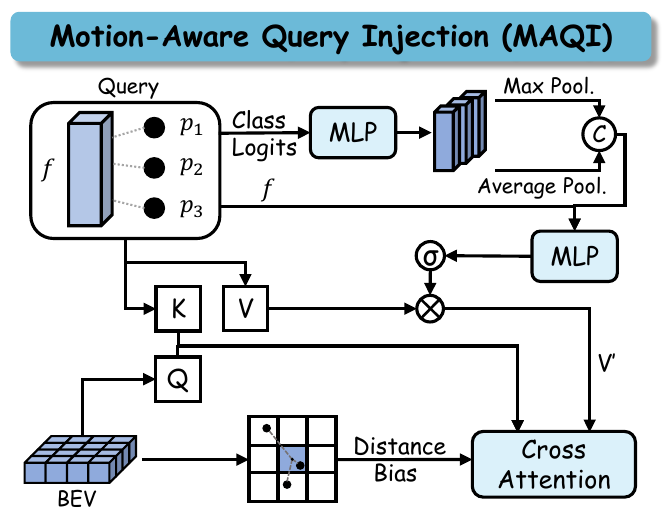}
    \caption{Motion-Aware Query Injection (MAQI). MAQI injects confidence-weighted query features into neighboring BEV tokens through distance-biased cross-attention.}
    \label{fig:maqi}
\end{wrapfigure}

Autoregressive propagation may introduce errors in query localization and
semantics. We therefore estimate a confidence score
$c_i^\tau\in[0,1]$ for each valid query before injection. As shown in
Figure~\ref{fig:maqi}, the point-wise class log-probabilities are embedded and
aggregated by average and max pooling into a semantic summary
$\mathbf s_i^\tau$. The confidence head combines this summary with the
normalized query feature:
\begin{equation}
    c_i^\tau =
    \sigma\!\left(
    \mathcal C\!\left[
    \operatorname{LN}(\mathbf f_i^\tau),
    \mathbf s_i^\tau
    \right]\right),
    \label{eq:query_confidence}
\end{equation}
where $\mathcal C$ is a lightweight MLP. The confidence target, defined in
Section~\ref{sec:objectives}, jointly considers localization and semantic
agreement.

At the final forecasting layer, the LST output
$\widetilde{\mathbf X}_\tau$ is reshaped into BEV tokens
$\mathbf B_\tau\in\mathbb R^{H_zW_z\times C}$, with
$\mathbf q_\tau^{\mathrm{bev}}=\mathbf B_\tau+\mathbf E_{\mathrm{bev}}$.
For each query, its feature, point-set centroid, relative point offsets, and
dynamic class distribution are combined into a query token, from which
$\mathbf k_i^\tau$ and $\mathbf v_i^\tau$ are generated. The BEV tokens then
cross-attend to the query keys and confidence-weighted values:
\begin{equation}
\mathbf B'_{\tau} =
\mathbf B_{\tau} +
\mathcal P_o\!\left(
\operatorname{MHA}
\left(
\mathbf q_{\tau}^{\mathrm{bev}},
\{\mathbf k_i^\tau\}_{i=1}^{N_q},
\{c_i^\tau \mathbf v_i^\tau\}_{i=1}^{N_q};
\mathbf b_\tau
\right)
\right),
    \label{eq:writeback}
\end{equation}

where $\mathbf b_\tau$ combines a Gaussian centroid-distance bias with the
invalid-query mask, and $\mathcal P_o$ is a zero-initialized output
projection. The distance bias localizes each query's influence to nearby BEV
tokens, while $c_i^\tau$ suppresses unreliable query updates. Finally, the
dense latent is updated through a zero-initialized residual predictor:
\begin{equation}
    \hat Z_\tau =
    \hat Z_{\tau-1} +
    \mathcal P_r(\operatorname{reshape}(\mathbf B'_\tau)).
    \label{eq:dense_residual_transition}
\end{equation}

\subsection{Learning Objectives}
\label{sec:objectives}

We supervise the dense rollout by aligning each predicted BEV latent with the
corresponding latent produced by the frozen tokenizer:
\begin{equation}
    \mathcal L_{\mathrm{latent}} =
    \sum_{k=1}^{F}\alpha_k\,\mathrm{MSE}
    (\hat Z_{t+k}, Z_{t+k}),
    \label{eq:latent_loss}
\end{equation}
where $\alpha_k\geq0$ weights forecast horizon $k$. To guide DQI toward moving
regions, the moving-mask head is supervised by a binary BEV target whose
positive cells belong to dynamic semantic classes and have flow magnitude above
a motion threshold:
\begin{equation}
    \mathcal L_{\mathrm{moving}} =
    \mathrm{BCE}(\hat{\mathbf M},\mathbf M)
    + \mathrm{Dice}(\hat{\mathbf M},\mathbf M),
    \label{eq:moving_loss}
\end{equation}
where $\hat{\mathbf M}$ and $\mathbf M$ denote the predicted probability map
and binary target, respectively. On InfraOcc, flow is generated offline from annotated 3D boxes and velocities as detailed in the appendix, and is used only after thresholding to identify moving occupancy cells. The model neither takes flow as input nor directly predicts the flow vectors.

For query supervision, moving BEV cells are converted into metric point sets
with semantic labels. For valid queries, $\mathcal L_{\mathrm{point}}$ uses a
class-weighted bidirectional nearest-neighbor Smooth-$L_1$ loss for point
alignment, while $\mathcal L_{\mathrm{cls}}$ applies class-balanced focal loss
to point semantics. Both losses supervise the pre-rollout refinement stages and
future query updates. We further supervise the query confidence used by MAQI.
Let $d_{ip}$ denote the metric distance from point $p$ of query $i$ to its
nearest ground-truth moving point, and $s_{ip}\in\{0,1\}$ indicates semantic
agreement. The confidence target is
\begin{equation}
    c_i^{*}=\frac{1}{N_p}\sum_{p=1}^{N_p}
    \exp\!\left(-\frac{d_{ip}^{2}}{2\sigma_r^{2}}\right)s_{ip},
    \label{eq:reliability_target}
\end{equation}
where $\sigma_r>0$ controls the localization tolerance, and the confidence loss
is $\mathcal L_{\mathrm{rel}}=\mathrm{BCE}(c_i,c_i^{*})$.

The overall training objective is
\begin{equation}
    \mathcal L = \lambda_l\mathcal L_{\mathrm{latent}}
    + \lambda_m\mathcal L_{\mathrm{moving}}
    + \lambda_p\mathcal L_{\mathrm{point}}
    + \lambda_c\mathcal L_{\mathrm{cls}}
    + \lambda_r\mathcal L_{\mathrm{rel}}.
    \label{eq:loss}
\end{equation}

\begin{table*}[t]
    \centering
    \caption{4D occupancy forecasting performance on InfraOcc validation set.}
    \label{tab:main_results}
    \begingroup
    \small
    \renewcommand{\arraystretch}{1.1}
    \setlength{\tabcolsep}{1.5pt}
    \newcolumntype{C}[1]{>{\centering\arraybackslash}m{#1}}
    \begin{tabular}{@{}C{4.15cm}|C{1.10cm}|*{4}{C{0.92cm}}|*{4}{C{0.92cm}}@{}}
        \toprule
        \multirow{2}{*}{Method} & \multirow{2}{*}{Input} &
        \multicolumn{4}{c|}{mIoU$_{\mathrm{dyn}}$ (\%) $\uparrow$} &
        \multicolumn{4}{c}{mIoU$_{\mathrm{sta}}$ (\%) $\uparrow$} \\
        \cline{3-6} \cline{7-10}
        & & 1\,s & 2\,s & 3\,s & Avg. & 1\,s & 2\,s & 3\,s & Avg. \\
        \midrule
        OccWorld-O (ECCV 2024) & 3D-Occ & 25.64 & 21.24 & 19.76 & 22.21 & \underline{97.52} & \underline{97.42} & \underline{97.37} & \underline{97.44} \\
        I$^2$-World-O (ICCV 2025) & 3D-Occ & \underline{33.34} & \underline{29.67} & \underline{28.66} & \underline{30.56} & 97.30 & 97.27 & 97.07 & 97.21 \\
        \rowcolor{blue!8} \textbf{InfraWorld-O (Ours)} & 3D-Occ & \textbf{46.03} & \textbf{38.08} & \textbf{34.63} & \textbf{39.58} & \textbf{97.79} & \textbf{97.72} & \textbf{97.68} & \textbf{97.73} \\
        \midrule
        OccWorld-STC (ECCV 2024) & Camera & 16.20 & 14.29 & 13.69 & 14.73 & 78.29 & 78.27 & 78.24 & 78.27 \\
        I$^2$-World-STC (ICCV 2025) & Camera & \underline{19.47} & \underline{16.95} & \underline{15.73} & \underline{17.38} & \underline{78.71} & \underline{78.70} & \textbf{78.69} & \underline{78.70} \\
        SparseWorld (AAAI 2026) & Camera & 8.52 & 8.28 & 8.16 & 8.32 & 73.67 & 73.80 & 73.12 & 73.53 \\
        \rowcolor{blue!8} \textbf{InfraWorld-STC (Ours)} & Camera & \textbf{22.52} & \textbf{20.73} & \textbf{20.43} & \textbf{21.23} & \textbf{78.75} & \textbf{78.71} & \underline{78.68} & \textbf{78.71} \\
        \bottomrule
    \end{tabular}
    \endgroup
    \vspace{-10pt}
\end{table*}

\vspace{-1.5\baselineskip}
\section{Experiments}
\label{sec:experiments}

\newcommand{\classswatch}[3]{%
    \begingroup
    \setlength{\fboxsep}{0pt}%
    \setlength{\fboxrule}{0.3pt}%
    \fcolorbox[RGB]{0,0,0}{#1,#2,#3}{\rule{0pt}{0.8em}\rule{0.8em}{0pt}}%
    \endgroup
}

\subsection{Experimental Setup}
\label{sec:experimental_setup}

\paragraph{Datasets and metrics.}
We evaluate our method on InfraOcc~\citep{InfraOcc} and Occ3D--nus~\citep{occ3d}. InfraOcc consists of 215 training sequences and 75 evaluation sequences collected in roadside scenarios, with occupancy grids of $320\times320\times16$ voxels at a resolution of 0.4\,m. We report mIoU$_{\mathrm{dyn}}$ and mIoU$_{\mathrm{sta}}$ over the dynamic and static categories, respectively. Occ3D--nus is a 4D occupancy forecasting benchmark built upon nuScenes, comprising 600 training scenes and 150 validation scenes. Its occupancy range spans $[-40,40]$\,m along both the $x$- and $y$-axes and $[-1,5.4]$\,m along the $z$-axis, with a voxel size of 0.4\,m and a sampling rate of 2\,Hz. On Occ3D--nus, we report mIoU at 1, 2, and 3\,s, together with the average over the three horizons.

\paragraph{Implementation details.}
We use 2 seconds of historical occupancy observations to forecast the subsequent 3 seconds. The tokenizer is pretrained for 24 epochs and subsequently frozen. InfraWorld-O is trained with ground truth 3D occupancy and InfraWorld-STC is trained with predictions from STCOcc~\citep{stcocc}. We train the model for 24 epochs on InfraOcc and 48 epochs on Occ3D--nus, using AdamW with an initial learning rate of $1\times10^{-3}$, a weight decay of $10^{-2}$, and a batch size of 8 per GPU on four NVIDIA RTX 4090 GPUs. All inference results are measured on a single NVIDIA RTX 4090 GPU with 24GB of memory.

\subsection{Main Results}
\label{sec:main_results}

\begin{wrapfigure}[14]{r}{0.62\linewidth}
    \vspace{-2.5\baselineskip}
    \centering
    \includegraphics[width=\linewidth]{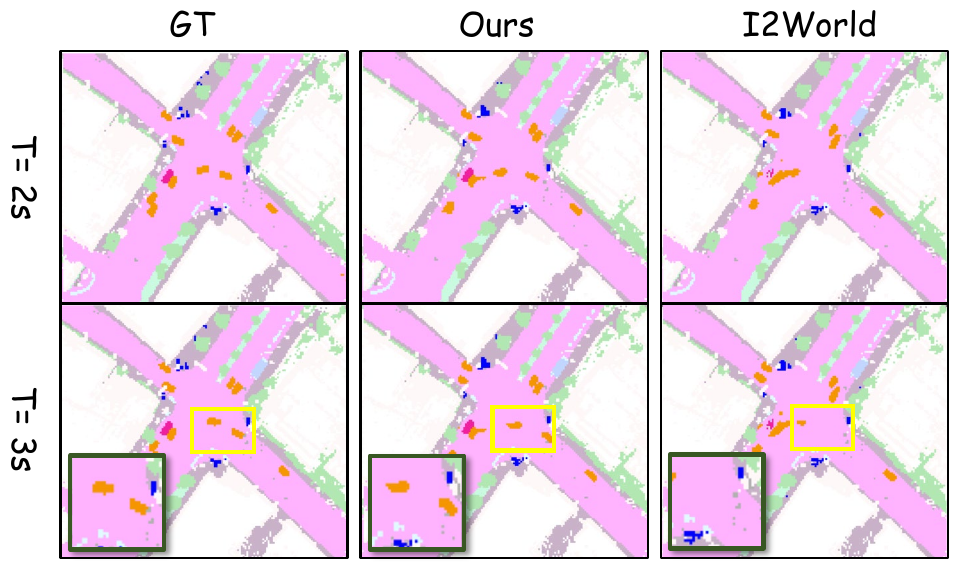}
    \caption{Qualitative comparison between InfraWorld and I$^2$-World on InfraOcc. Yellow boxes highlight dynamic regions where InfraWorld more closely matches the ground truth.}
    \label{fig:qualitative_forecasting}
    \vspace{-0.8\baselineskip}
\end{wrapfigure}

\begingroup
\setlength{\columnsep}{4pt}
\begin{wraptable}[14]{r}{0.60\linewidth}
    \setlength{\intextsep}{0pt}
    \vspace{-4\baselineskip}
    \centering
    \caption{Forecasting performance on Occ3D--nus.}
    \label{tab:comparison_results}
    \begingroup
    \small
    \setlength{\tabcolsep}{0pt}
    \renewcommand{\arraystretch}{1.1}
    \newcolumntype{C}[1]{>{\centering\arraybackslash}m{#1}}
    \newcolumntype{Y}{>{\centering\arraybackslash}X}
    \begin{tabularx}{\linewidth}{@{}C{2.31cm}|C{1.95cm}|*{4}{Y}@{}}
        \toprule
        \multirow{2}{*}{Method} & \multirow{2}{*}{Input} &
        \multicolumn{4}{c}{mIoU (\%) $\uparrow$} \\
        \cline{3-6}
        & & 1\,s & 2\,s & 3\,s & Avg. \\
        \midrule
        OccWorld-O & Occ & 25.78 & 15.14 & 10.51 & 17.14 \\
        OccLLaMA-O & Occ \& Text & 25.05 & 19.49 & 15.26 & 19.93 \\
        DFIT-OccWorld & Occ \& Camera & 31.68 & 21.29 & 15.19 & 22.71 \\
        DOME & Occ & 35.11 & 25.89 & 20.29 & 27.10 \\
        UniScene & Occ & 35.37 & 29.59 & 25.08 & 31.76 \\
        I$^2$-World-O & Occ & \underline{47.62} & \underline{38.58} & \underline{32.98} & \underline{39.73} \\
        \rowcolor{blue!8} \textbf
        {InfraWorld-O} & Occ & \textbf{50.34} & \textbf{41.16} & \textbf{35.25} & \textbf{42.25 } \\
        \midrule
        OccWorld-STC & Camera & 15.16 & 10.97 & 10.19 & 12.11 \\
        SparseWorld & Camera & 14.93 & 13.15 & 11.51 & 13.20 \\
        I$^2$-World-STC & Camera & \underline{19.60} & \underline{16.71} & \underline{14.48} & \underline{16.93} \\
        \rowcolor{blue!8} \textbf{InfraWorld-STC} & Camera & \textbf{20.56} & \textbf{18.50} & \textbf{16.61} & \textbf{18.56} \\
        \bottomrule
    \end{tabularx}
    \endgroup
\end{wraptable}

Table~\ref{tab:main_results} presents the main comparison on InfraOcc. With occupancy inputs, InfraWorld-O achieves an average mIoU$_{\mathrm{dyn}}$ of 39.58\%, outperforming I$^2$-World-O by 9.02 points, and retains a 5.97-point advantage at the 3\,s horizon. This improvement validates our key insight that persistent motion queries can complement the dense scene state by providing dedicated modeling capacity for the sparse localized regions that drive future evolution. Meanwhile, InfraWorld-O maintains a high average mIoU$_{\mathrm{sta}}$ of 97.73\%, indicating that the dense pathway effectively preserves the stable scene layout while the query pathway focuses on evolving regions.

Table~\ref{tab:comparison_results} further evaluates InfraWorld on Occ3D--nus, an ego-centric benchmark with scene dynamics distinct from the roadside setting. InfraWorld-O and InfraWorld-STC achieve average mIoUs of 42.25\% and 18.56\%, respectively, consistently outperforming their I$^2$-World counterparts across all forecasting horizons by 2.52 and 1.63 points on average. These results show that the proposed dense scene state and focused motion reasoning remain effective for ego-centric occupancy forecasting.

Figure~\ref{fig:qualitative_forecasting} provides a qualitative comparison with I$^2$-World. Both methods largely preserve the static roadside layout, whereas the differences are more evident in localized regions containing moving participants. At the longer 3\,s horizon, I$^2$-World tends to miss or distort the occupancy of moving objects, while InfraWorld better preserves their spatial locations and shapes, producing predictions that more closely match the ground truth in the highlighted regions.
This behavior is consistent with our design, where LST integrates recent temporal continuity and longer-term motion trends to provide more informative temporal cues, while persistent motion queries recurrently refine localized motion states throughout the rollout.

\par
\endgroup

Together with the dense scene state that preserves the stable roadside context, these components enable more accurate prediction of localized traffic evolution over longer horizons.

\subsection{Ablation Studies}
\label{sec:ablations}

\paragraph{Overall ablation.}

Table~\ref{tab:component_ablation} evaluates the contribution of the principal components of InfraWorld. For variants with MAQI but without DQI, we initialize one query at the center of every tile. Starting from the dense forecasting baseline, introducing LST improves the average mIoU$_{\mathrm{dyn}}$ by 1.94 points. By combining recent-frame continuity with accumulated changes over the observed history, LST captures both the immediate state of local motion and its longer-term evolution trend, providing stronger temporal cues for forecasting where and how the scene will change. MAQI produces a larger gain of 3.42 points, supporting our central design that persistent motion queries provide complementary information for localized regions whose dynamics are difficult to capture through the dense state alone. When DQI is further introduced, mIoU$_{\mathrm{dyn}}$ increases by another 2.40 points, indicating that assigning the limited query budget according to observed motion evidence makes this localized reasoning more effective. Combining all three components yields 39.58\% mIoU$_{\mathrm{dyn}}$, an 8.65-point improvement over the baseline. These results validate that preserving a complete dense scene state while allocating dedicated modeling capacity to sparse evolving regions is effective for roadside occupancy forecasting.

\begingroup
\setlength{\columnsep}{4pt}
\begin{wraptable}[10]{r}{0.5\linewidth}
    \setlength{\intextsep}{0pt}
    \vspace{-2\baselineskip}
    \centering
    \caption{Component-wise ablation of InfraWorld.}
    \label{tab:component_ablation}
    \begingroup
    \small
    \setlength{\tabcolsep}{1pt}
    \renewcommand{\arraystretch}{1.1}
    \newcolumntype{Y}{>{\centering\arraybackslash}X}
    \begin{tabularx}{\linewidth}{@{}>{\centering\arraybackslash}m{2.6em}|>{\centering\arraybackslash}m{2.6em}|>{\centering\arraybackslash}m{2.6em}|*{4}{Y}@{}}
        \toprule
        \multirow{2}{*}{LST} & \multirow{2}{*}{MAQI} & \multirow{2}{*}{DQI} &
        \multicolumn{4}{c}{mIoU$_{\mathrm{dyn}}$ (\%) $\uparrow$} \\
        \cline{4-7}
        & & & 1\,s & 2\,s & 3\,s & Avg. \\
        \midrule
        -- & -- & -- & 33.68 & 32.54 & 26.58 & 30.93 \\
        \boldmath$\checkmark$ & -- & -- & 35.20 & 32.95 & 30.46 & 32.87 \\
        -- & \boldmath$\checkmark$ & -- & 36.81 & 34.33 & 31.91 & 34.35 \\
        -- & \boldmath$\checkmark$ & \boldmath$\checkmark$ & 42.76 & 34.87 & 32.62 & 36.75 \\
        \rowcolor{cyan!8}
        \boldmath$\checkmark$ & \boldmath$\checkmark$ & \boldmath$\checkmark$ & \textbf{46.03} & \textbf{38.08} & \textbf{34.63} & \textbf{39.58} \\
        \bottomrule
    \end{tabularx}
    \endgroup
\end{wraptable}

\paragraph{Ablation on LST.}
Table~\ref{tab:lst_maqi_ablation} examines the two temporal branches in LST. Removing either the short-term or long-term branch degrades performance, with drops of 1.33 and 1.48 points in average mIoU$_{\mathrm{dyn}}$, respectively. The short-term branch exploits the strong temporal continuity between adjacent frames under the fixed roadside coordinate system, while the long-term branch summarizes residual changes across the history to capture accumulated motion trends. Their comparable contributions indicate that recent temporal continuity and longer-term motion trends capture complementary aspects of scene evolution, and jointly provide a richer temporal representation for forecasting dynamic regions. Therefore, their combination provides a more suitable temporal representation for both dense-state prediction and subsequent query refinement.
\par
\endgroup

\begin{figure*}[!t]
    \centering
    \includegraphics[width=\textwidth]{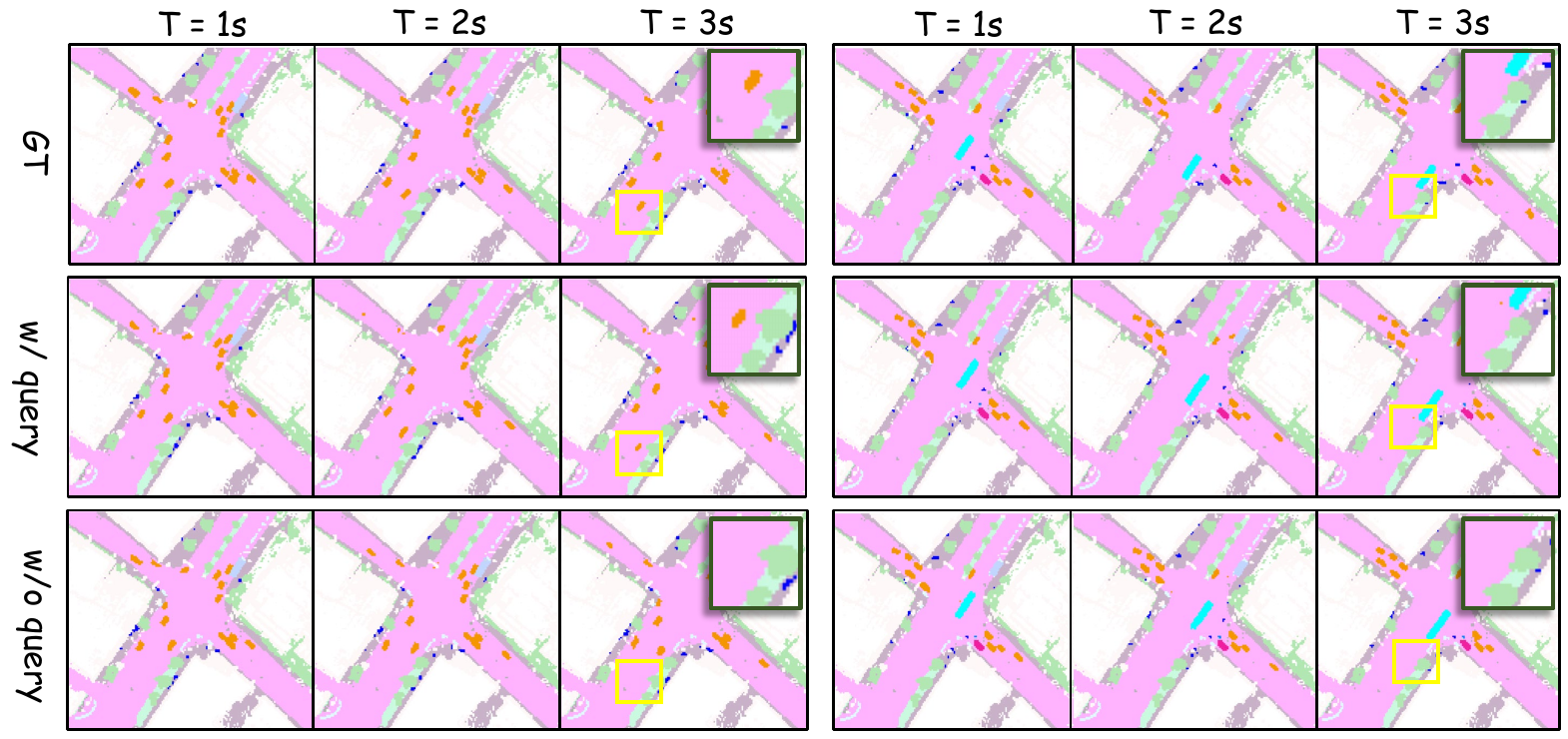}
    \caption{Qualitative comparison of occupancy forecasts with and without dynamic queries. Yellow boxes highlight moving regions where dynamic queries improve agreement with the GT.}
    \label{fig:query_forecasting_ablation}
    \vspace{-10pt}
\end{figure*}

\paragraph{Ablation on MAQI.}
We further compared different strategies for writing query information into the dense BEV state. Replacing MAQI with bilinear splatting reduces the average mIoU$_{\mathrm{dyn}}$ from 39.58\% to 37.78\%. Unlike fixed spatial interpolation, the distance-biased cross-attention in MAQI allows each query to interact with nearby BEV features according to the local scene context, providing a more adaptive mechanism for localized updates. Removing confidence weighting further decreases performance to 38.18\%. This confirms the importance of filtering query updates during autoregressive rollout, where localization and semantic errors can accumulate over time. By suppressing unreliable writeback, MAQI allows the query pathway to refine evolving regions without unnecessarily disturbing the dense scene state. The qualitative results in Figure~\ref{fig:query_forecasting_ablation} show that introducing dynamic queries produces more accurate forecasts of moving objects. This improvement is consistent with persistent queries maintaining and recurrently updating localized motion states throughout the rollout, providing dedicated motion cues for predicting future occupancy changes in these localized regions.

\begin{table*}[t]
    \centering
    \caption{Joint ablation of LST and MAQI on InfraOcc.}
    \label{tab:lst_maqi_ablation}
    \begingroup
    \small
    \setlength{\tabcolsep}{3pt}
    \renewcommand{\arraystretch}{1.1}
    \newcolumntype{Y}{>{\centering\arraybackslash}X}
    \begin{tabularx}{\textwidth}{@{}>{\centering\arraybackslash}m{12em}|>{\centering\arraybackslash}m{10em}|*{4}{Y}@{}}
        \toprule
        \multirow{2}{*}{Fusion method} & \multirow{2}{*}{Writeback method} &
        \multicolumn{4}{c}{mIoU$_{\mathrm{dyn}}$ (\%) $\uparrow$} \\
        \cline{3-6}
        & & 1\,s & 2\,s & 3\,s & Avg. \\
        \midrule
        w/o short-term fusion & MAQI & 44.06 & 36.67 & 34.03 & 38.25 \\
        w/o long-term fusion & MAQI & 45.02 & 36.18 & 33.09 & 38.10 \\
        \rowcolor{cyan!8}
        LST & MAQI & \textbf{46.03} & \textbf{38.08} & \textbf{34.63} & \textbf{39.58} \\
        LST & Bilinear Splat & 42.83 & 36.18 & 34.34 & 37.78 \\
        LST & MAQI w/o conf. & 45.11 & 36.32 & 33.11 & 38.18 \\
        \bottomrule
    \end{tabularx}
    \endgroup
    \vspace{-10pt}
\end{table*}

\subsection{Model Efficiency}
\label{sec:model_efficiency}

\begingroup
\setlength{\columnsep}{4pt}
\begin{wraptable}[7]{r}{0.61\linewidth}
    \setlength{\intextsep}{0pt}
    \vspace{-3.5\baselineskip}
    \centering
    \caption{Efficiency--accuracy comparison. Lat: bs=1 inference latency.}
    \label{tab:efficiency}
    \begingroup
    \small
    \setlength{\tabcolsep}{1pt}
    \renewcommand{\arraystretch}{1.1}
    \newcolumntype{C}[1]{>{\centering\arraybackslash}m{#1}}
    \begin{tabular*}{\linewidth}{@{\extracolsep{\fill}}C{2.00cm}|C{1.70cm}|C{1.55cm}|C{1.25cm}|C{1.60cm}@{}}
        \toprule
        Method & \mbox{Params. (M)} & \mbox{FLOPs (G)} & \mbox{Lat. (ms)} &
        \mbox{mIoU$_{\mathrm{dyn}}$ $\uparrow$} \\
        \midrule
        OccWorld & 71.1 & 1291 & 162 & 22.21 \\
        I$^2$-World & 22.5 & 91 & 85 & 30.56 \\
        I$^2$-World-L & 28.9 & 169 & 91 & 32.90 \\
        \rowcolor{blue!8} InfraWorld & 29.1 & 166 & 119 & 39.58 \\
        \bottomrule
    \end{tabular*}
    \endgroup
\end{wraptable}

Table~\ref{tab:efficiency} compares parameters, FLOPs, inference latency and forecasting accuracy. InfraWorld introduces a moderate increase in
computation over I$^2$-World while improving mIoU$_{\mathrm{dyn}}$ by 9.02
points. To isolate the effect of model capacity, we further enlarge
I$^2$-World to I$^2$-World-L with a parameter count comparable to InfraWorld.
Despite similar model size and FLOPs, I$^2$-World-L still trails
InfraWorld by 6.68 points, indicating that the improvement does not simply
arise from increased model capacity but from the targeted designs for roadside
scene evolution. DQI allocates persistent queries to
motion-supported regions, LST integrates recent continuity with longer-term
motion trends, and MAQI injects reliable query-level dynamics into the
corresponding local BEV regions. InfraWorld also remains more
efficient than OccWorld, achieving a favorable accuracy--efficiency trade-off.
\par
\endgroup

\section{Conclusion}

In this work, we formulate roadside occupancy forecasting as selective state evolution and propose \textbf{InfraWorld}, which maintains a dense BEV state for complete scene context while using persistent motion queries to model sparse localized dynamics. DQI allocates queries to motion-supported regions, LST integrates recent continuity with longer-term motion trends, and MAQI selectively injects reliable query-level dynamics into local BEV regions. Together, these designs preserve the stable roadside layout while concentrating predictive capacity on the regions that drive future evolution. Experiments on InfraOcc and Occ3D--nus validate the effectiveness of InfraWorld for roadside forecasting and demonstrate its broader applicability to ego-centric settings.

\subsection*{AI use statement}

In this work, we used generative AI tools to assist with code development, including drafting and debugging auxiliary code, and to improve the presentation of the manuscript, including language polishing and table formatting. All AI-assisted content was carefully reviewed by the authors. AI-assisted code was inspected and tested for correctness before being used in our experiments. Language revisions were manually checked to ensure that they preserved the intended technical meaning and did not alter the scientific claims. AI-assisted table formatting was verified against the original experimental results to ensure numerical accuracy and consistency. We take responsibility for the final content of this work, including text, claims, code, and artifacts produced with the aid of generative AI.

\bibliography{iclr2027}
\bibliographystyle{iclr2027}

\end{document}